\documentclass{article}

\usepackage{arxiv}

\usepackage[utf8]{inputenc} 
\usepackage[T1]{fontenc}    
\usepackage{hyperref}       
\usepackage{url}            
\usepackage{booktabs}       
\usepackage{amsfonts}       
\usepackage{nicefrac}       
\usepackage{microtype}      
\usepackage{lipsum}		
\usepackage{graphicx}
\usepackage{natbib}
\usepackage{doi}
\usepackage{xcolor}
\usepackage{listings}

\title{Large Language Models for Failure Mode Classification: an Investigation}

\author{ \href{https://orcid.org/0000-0001-6494-7015}{\includegraphics[scale=0.06]{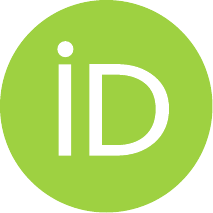}\hspace{1mm}Michael~Stewart} \\
	Department of Computer Science and Software Engineering\\
	The University of Western Australia\\
	Perth, Western Australia \\
	\texttt{michael.stewart@uwa.edu.au} \\
	\And
	\href{https://orcid.org/0000-0002-7336-3932}{\includegraphics[scale=0.06]{orcid.pdf}\hspace{1mm}Melinda Hodkiewicz} \\
	School of Engineering\\
	The University of Western Australia\\
	Perth, Western Australia \\
	\texttt{melinda.hodkiewicz@uwa.edu.au} \\
    \And
	\href{https://orcid.org/0000-0002-2504-3790}{\includegraphics[scale=0.06]{orcid.pdf}\hspace{1mm}Sirui Li} \\
	Department of Computer Science and Software Engineering\\
	The University of Western Australia\\
	Perth, Western Australia \\
	\texttt{sirui.li@uwa.edu.au} \\
}

\renewcommand{\shorttitle}{LLMs for FMC: An Investigation}

\hypersetup{
pdftitle={GPT FMC},
pdfsubject={q-bio.NC, q-bio.QM},
pdfauthor={Michael Stewart, Sirui Li, Melinda Hodkiewicz},
pdfkeywords={Technical Language Processing, Failure Mode Classification, Large Language Models},
}

\begin{document}
\maketitle

\begin{abstract}
   In this paper we present the first investigation into the effectiveness of Large Language Models (LLMs) for Failure Mode Classification (FMC). FMC, the task of automatically labelling an observation with a corresponding failure mode code, is a critical task in the maintenance domain as it reduces the need for reliability engineers to spend their time manually analysing work orders. We detail our approach to prompt engineering to enable an LLM to predict the failure mode of a given observation using a restricted code list. We demonstrate that the performance of a GPT-3.5 model (F1=0.80) fine-tuned on annotated data is a significant improvement over a currently available text classification model (F1=0.60) trained on the same annotated data set. The fine-tuned model also outperforms the out-of-the box GPT-3.5 (F1=0.46). This investigation reinforces the need for high quality fine-tuning data sets for domain-specific tasks using LLMs. 
\end{abstract}


\section{Introduction}

The maintenance of assets plays a critical role in the safety and costs of industrial organisations. One of the key tasks within maintenance is failure mode identification. This task is done by reliability engineers to capture and code failure and other undesirable events. These failure mode codes, together with data such as the cost/ production/ service impact, safety and environmental consequence of the event are used to prioritise improvement work, update maintenance strategy and can assist product/ plant engineers to improve future design by updating their failure modes and effects analysis.  Consistent and reproducible failure mode code assignment is  difficult as the observation of each event are captured  by field technicians in natural language. For example, consider the following maintenance work order texts:

\begin{itemize}
    \item \texttt{pump runs for a while and trip}
    \item \texttt{engin does not work}
    \item \texttt{pmp spraying out slurry}
    \item \texttt{seal leaking}
    \item \texttt{leak in seal}
\end{itemize}

Each of these work orders contain an observation made by the field technician, such as ``does not work'', ``leaking'', and so on. In any maintenance management system there are thousands of these observations and each needs a failure mode classification (FMC), such as ``leaking'' and ``breakdown''  according to an agreed list. The challenge, is that each person doing the coding, whether it be the technician generating the work order, or the reliability engineer reviewing it, comes with their own mental model of the asset and its behaviour \citep{sexton2019categorization}. Further, attention to the task of coding accurately is influenced by factors such as training, managerial support, technological input control and motivation \citep{murphy2009improving, unsworth2011goal,molina2013managerial}. It is too expensive to have university-trained reliability engineers review each of these codes manually given the volume. The opportunity for AI to assist in failure mode classification is therefore an active research area \citep{sexton2018benchmarking,akhbardeh2020nlp,sala2022nlp,stewart2022mwo2kg,usuga2022using}. 


There has recently been a surge of interest in Large Language Models (LLMs), predominately as the result of the popularity of chatbot interfaces such as ChatGPT\footnote{\url{https://chat.openai.com/}}. LLMs such as OpenAI's GPT-3.5\footnote{\url{https://platform.openai.com/docs/models}} have been trained on massive corpora and thus encapsulate knowledge from a wide variety of domains. It has also been shown that LLMs require little to no fine-tuning, meaning they exhibit excellent performance with barely any annotated training data~\citep{brown2020language}. Rather than focusing on developing manually-annotated datasets to train models (like with more ``traditional'' text classification models such as Flair~\citep{akbik2018coling}), users of LLMs typically employ \textit{prompt engineering} in order to craft their input prompt to elicit a particular response from the model. As a result of their excellent performance on a wide range of natural language processing tasks, LLMs have already been applied to a variety of domains. Examples include medicine~\citep{singhal2022large, thirunavukarasu2023large}, 
education~\citep{kasneci2023chatgpt}, and 
vehicle accident records~\citep{mumtarin2023large}. 

However, to the best of our knowledge, no research has yet investigated the use of LLMs within the maintenance domain, let alone specifically for FMC. In light of this research gap, and the potential for automated FMC to enable significant time and cost benefits to industry, we present an investigation into the effectiveness of using Large Language Models for Failure Mode Classification. Our contributions are as follows:

\begin{itemize}
    \item We investigate the most effective prompt format for performing FMC using an LLM without any fine-tuning.
    \item We determine whether it is necessary to fine-tune an LLM on a set of annotated data to achieve good FMC performance.
    \item We provide a comparison between the performance of fine-tuned LLMs and text classification models for FMC.
\end{itemize}

This paper is structured as follows. We begin by providing an outline of our models, methods and experiments, and detail the dataset that we use for fine-tuning and evaluation. We then present our results, which directly tie in to our contributions above. Finally, we present our conclusion and an outlook to future work.

The source code of this paper is open source and is available on \href{https://github.com/nlp-tlp/chatgpt-fmc}{GitHub}.

\section{Methods}

The aim of this paper is to evaluate the applicability of Large Language Models (LLMs) to Failure Mode Classification (FMC). In this section we provide an overview of the dataset we are using for our evaluation, as well as the models that we evaluate in Section~\ref{sec:results}.

\subsection{Dataset}

The dataset on which we evaluate each model is an extract from the annotated maintenance work order dataset introduced by~\citep{stewart2022mwo2kg} and available on PapersWithCode\footnote{\url{https://paperswithcode.com/dataset/fmc-mwo2kg}}. The data set consists of 502 (observation, label) pairs for training, 62 for validation, and 62 for testing. The observations, which are written in natural language, were extracted from a set of maintenance work orders using Named Entity Recognition (NER). The labels are taken from a set of 22 failure mode codes from ISO 14224~\footnote{\url{https://www.iso.org/standard/64076.html}}. Each observation was labelled by a domain expert. Some examples from this dataset are as follows:

\begin{itemize}
    \item broken, Breakdown
    \item leaking fluid, Leaking
    \item too hot, Overheating
    \item triping, Electrical
    \item not starting, Failure to start on demand
\end{itemize}

This open data set and the model presented in ~\citep{stewart2022mwo2kg} represent the state-of-the-art for FMC in the literature at this point in time and hence are used for comparative purposes.

\subsection{Models}

We evaluate the following models:

\begin{enumerate}
    \item \textbf{Flair}: A Flair-based~\citep{akbik2018coling} text classification model, trained on the annotated dataset.
    \item \textbf{GPT-3.5}: The off-the-shelf \texttt{GPT-3.5-Turbo} model from OpenAI.
    \item \textbf{GPT-3.5 (Fine-tuned)}: The \texttt{GPT-3.5-Turbo} model, fine-tuned on the annotated dataset.
\end{enumerate}

The Flair model is a Bidirectional Long Short-Term Memory-based~\citep{hochreiter1997long} text classification model that takes a sequence of text as input, and predicts a single label. This is the same model as used in~\citep{stewart2022mwo2kg}, and further implementation details are available in the respective paper. 

The first layer of the model, the embedding layer, was pre-trained by the Flair developers on a corpora of web, Wikipedia data, and subtitles, and thus the model has little innate knowledge of maintenance. The model was trained by~\citep{stewart2022mwo2kg} on the dataset of 502 (observation, label) pairs and validated on the 62-pair validation set. In contrast to the GPT-based models, the computational requirements of training and using this model are low enough to be able to train on most desktop computers. This also means it can be used offline, and is thus appropriate for handling sensitive data. 

The LLM-based models are based on OpenAI's GPT-3.5~\citep{brown2020language}\footnote{GPT-4.0 was not available for fine-tuning as of the time of writing, hence the decision to use GPT-3.5.}, the model behind  ChatGPT\footnote{\url{https://chat.openai.com/}}. The \texttt{GPT-3.5} model is ``off-the-shelf'' in that we are using the model without any form of fine-tuning. We are relying on the model's knowledge of maintenance that it has gleaned from its massive training corpora in order to task it to perform failure mode classification. The \texttt{GPT-3.5 (Fine-tuned)} model, on the other hand, is fine-tuned on the annotated dataset of 502 (observation, label) pairs, and validated on the 62-pair validation set.





\subsection{Data preparation}
\lstset{basicstyle=\footnotesize\ttfamily,breaklines=true}

\begin{lstlisting}[caption={An example prompt that is fed into the \texttt{GPT-3.5} and \texttt{GPT-3.5 (Fine-tuned)} models. The role of the \texttt{assistant} is only used during fine-tuning. \label{prompt-example}},captionpos=b]
    [{
        "role": "system",
        "content": "Determine the failure mode of the observation
                    provided by the user."
    },
    {
        "role": "user",
        "content": "too hot"
    },
    {
        "role": "assistant",
        "content": "Overheating"
    }]

\end{lstlisting}

The default behaviour of the GPT-based models is to act as a chatbot, and thus it will not respond with a failure mode code for a given observation unless the instruction to do so is included as part of the prompt. Structuring an input prompt to elicit a particular response from a large language model is known as \textit{prompt engineering}.

The latest versions of the GPT-based models require a three-part prompt. The \texttt{system}-level prompt dictates the desired response format of the model. For example, one can use this prompt to ask the model to reply in a sarcastic tone, or to reply with a one-word answer, and so on. The \texttt{user}-level prompt is the input from the user. Finally, the \texttt{assistant}-level prompt is the desired input from the LLM (this is used when fine-tuning to inform the model of the expected output). 

To create the prompts, we wrote Python code to iterate through the annotated CSV-based dataset and convert each (observation, label) pair into a prompt as shown in Listing 1. The same \texttt{system}-level prompt is used for each input to the model, and describes the task to perform (failure mode classification). We use the \texttt{user}-level prompt to provide the model with the observation that we want it to label. During the fine-tuning of the \texttt{GPT-3.5 (Fine-tuned)}, we include an \texttt{assistant}-level prompt that informs the model of the desired output for each observation (i.e. the failure mode). The design behind these prompts were based on the best practices listed in the OpenAI Documentation\footnote{\url{https://platform.openai.com/docs/guides/fine-tuning/fine-tuning-examples}}.

In our experiments we also investigate the necessity to add the following two texts to the \texttt{system}-level prompt:
\begin{itemize}
    \item In Section~\ref{subsec:results1}, we include the sentence ``Your answer should contain only the failure mode and nothing else.'' to instruct the language model to avoid outputting unnecessary text (e.g. ``The failure mode is ...'', etc.
    \item In Section~\ref{subsec:results2} we include ``Valid failure modes are: '' followed by a newline-separated list of valid labels from the dataset. This is an attempt to ensure that the model does not come up with its own failure modes, but instead outputs a failure mode code from the prescribed list.
\end{itemize}

\subsection{Evaluation metrics}

In the same manner as~\citep{stewart2022mwo2kg}, we evaluate each model using Micro F1 and Macro F1 score. Micro F1 calculates an F1-Score by adding the true positives (TPs), false positives (FPs) and false negatives (FNs) from all class labels together and then calculating F1-Score:

\begin{equation}
    Micro F1 = F1_{(class 1 + class 2 + ... + class n)}
\end{equation}

Macro f1, on the other hand, simply averages the F1-Score of each class. Given N is the number of class labels, it is calculated as follows:

\begin{equation}
    Macro F1 = \frac{\sum_{n \in N} F1_{class_n}  }{N}
\end{equation}

\section{Results}
\label{sec:results}

This section aims to answer the following questions:

\begin{enumerate}
    
    \item How best to use an off-the-shelf Large Language Model (LLM) to perform Failure Mode Classification (FMC)?
    \item Is it necessary to fine-tune the LLM to perform FMC?
    \item Are LLMs more effective at FMC than text classification models?
    \item What are some barriers one may face when using LLMs for FMC?
\end{enumerate}

\subsection{How best to use an off-the-shelf LLM to perform Failure Mode Classification?}
\label{subsec:results1}

\begin{table}[]
    \centering
    \begin{tabular}{l|p{12cm}}
        \textbf{Observation} & \textbf{LLM output} \\ \hline
runs for a while and trip & The failure mode of the observation "runs for a while and trips" suggests an electrical failure. This could be due to an overload, short circuit, or other electrical issue that causes the equipment to shut down or trip a circuit breaker. \\
    does not work & The failure mode of the observation "does not work" is a breakdown. \\
    spilling & The failure mode of the observation "spillage" is a leakage. \\
    spraying out slurry & The failure mode of the observation "spraying out slurry" is leaking. \\ \hline
    \end{tabular}
    \caption{Some examples of predictions made by the off-the-shelf \texttt{GPT-3.5-Turbo} on a sample of the test data. The \texttt{system}-level prompt is ``Determine the failure mode of the observation provided by the user.''}
    \label{tab:fmc-no-finetune-no-constraint}
\end{table}

\begin{table}[]
    \centering
    \begin{tabular}{l|p{12cm}}
        \textbf{Observation} & \textbf{LLM output} \\ \hline
runs for a while and trip & Overheating \\
very stiff to operate & Stiff operation \\
requires rebuild &Noisy operation \\
has no equipment earth & N/A \\
high earth reading & No failure mode can be determined from the given observation. \\
failed electrical & Failure mode: Electrical failure \\ \hline
    \end{tabular}
    \caption{Some examples of predictions made by the off-the-shelf \texttt{GPT-3.5-Turbo} on a sample of the test data. The \texttt{system}-level prompt is ``Determine the failure mode of the observation provided by the user. Your answer should contain only the failure mode and nothing else.''}
    \label{tab:fmc-no-finetune}
\end{table}


To address the first research question we begin by investigating the use of a simple \texttt{system}-level prompt of ``Determine the failure mode of the observation provided by the user.''. Upon feeding this prompt into the model, along with the \texttt{user}-level prompt (the observation, e.g. ``runs for a while and trip''), the LLM produces outputs as shown in Table~\ref{tab:fmc-no-finetune-no-constraint}. These outputs, which are conversational in nature, are not machine-readable and are therefore not applicable to downstream analysis. A more specific prompt is needed to perform FMC.


In light of this, we next add the phrase ``Your answer should contain only the failure mode and nothing else.'' to the \texttt{system}-level prompt. Adding this sentence to the prompt results in the model predicting a single failure mode for each observation, as shown in Table~\ref{tab:fmc-no-finetune}. However, there are several notable issues with the outputs of the model after adding this phrase. Firstly, despite the addition of the phrase in the prompt, the model still occasionally adds additional text to its response. One such example is its response for the phrase ``failed electrical'', to which it also adds ``Failure mode: '' prior to the actual classification. It also occasionally disregards the instruction when it was not capable of recognising a particular failure mode, for example in its classification of ``high earth reading''.

While the LLM is capable of predicting failure modes using this prompt, they are not aligned with any particular failure mode ontology. Downstream analysis using these failure modes is thus not possible, due to the sheer number of possible failure modes and inconsistency between them. For example, the model predicts both ``Leakage'' and ``Leaking'', which are the same failure mode written two different ways. One can liken the LLM's predicted failure modes to that which might be produced by a layperson, i.e. not a domain expert.


The non fine-tuned model also has difficulties producing consistent failure mode labels when dealing with uncertainty. When the model is unable to classify the observation, it responds in a variety of different ways, for example ``Insufficient information'', ``N/A'', ``None'', ``No failure mode detected.'', ``No failure mode provided.'', and so on. Attempting to resolve all possible variations of these phrases into a single classification (such as ``Unknown'' or ``Other'') is a non-trivial task, and thus the outputs of this model are not readily applicable to downstream tasks.


In an attempt to solve this issue we add a final phrase to the prompt: ``Valid failure modes include: '' followed by a newline-separated list of the failure mode labels appearing across the entire dataset. We found that this addition generally causes the model to behave as expected. However, it occasionally hallucinates labels: for example, it predicts the label ``Fail to open'' for ``sticking shu'', and ``Fail to adjust'' for ``cant be adjusted''. It also has issues with label consistency - for example, it predicts both ``Fail to function'' and ``Failure to function''. Similarly to the previous attempt without constraining the label space, this attempt at using the LLM directly without fine-tuning is not directly applicable to failure mode analysis as a result of these issues.

In summary we have demonstrated that it is possible to engineer the prompt to enable the LLM to predict failure mode codes without any fine-tuning. However, these outputs are not grounded in any particular ontology and are inconsistent. 



\subsection{Is it necessary to fine-tune the LLM to perform Failure Mode Classification?}

\begin{table*}[t]
\centering
\begin{tabular}{l|c|c|c|c}
\textbf{Failure mode} &
\textbf{Support} &
\textbf{Flair} &
\textbf{GPT-3.5} &
\textbf{GPT-3.5 (FT)} 
\\ \hline

Abnormal instrument reading & 1  &   1.00 & 1.00 & 0.00  \\ \hline
                  Breakdown & 7  &   0.37 & 0.44 & \textbf{1.00}  \\ \hline
              Contamination & 1  &   1.00 & 1.00 & 1.00 \\ \hline
                 Electrical & 6  &   0.67 & 0.50 & 0.67\\ \hline 
             Erratic output & 1  &   0.00 & 0.00 & 0.00 \\ \hline
           Fail to function & 3  &   \textbf{0.50} & 0.00 & 0.00  \\ \hline
 Failure to start on demand & 1  &   0.40 & 0.33 & \textbf{1.00}  \\ \hline
  Failure to stop on demand & 1  &   0.00 & 1.00 & \textbf{1.00} \\ \hline
                High output & 1  &   0.00 & 1.00 & 1.00 \\ \hline
                    Leaking & 3  &   0.67 & 0.86 & 1.00\\ \hline
                 Low output & 2  &   0.00 & 0.00 & 0.00 \\ \hline        
  Minor in-service problems & 17 &   0.73 & 0.11 & \textbf{1.00} \\ \hline
                      Other & 2  &   \textbf{0.67} & 0.40 & 0.00 \\ \hline  
                Overheating & 4  &   1.00 & 1.00 & 1.00 \\ \hline
           Plugged / choked & 6  &   0.67 & 0.25 & \textbf{1.00} \\ \hline
              Spurious stop & 1  &   0.00 & 0.00 & 0.00 \\ \hline
      Structural deficiency & 3  &   0.60 & 0.57 & \textbf{1.00} \\ \hline
                  Vibration & 2  &   0.67 & 1.00 & 1.00 \\ \hline

\multicolumn{2}{l}{\textbf{Micro-F1}} & 0.60 & 0.46 & \textbf{0.81}  \\ \hline 
\multicolumn{2}{l}{\textbf{Macro-F1}} & 0.46 & 0.53 & \textbf{0.62}  \\ \hline 
 
\end{tabular}

\caption{A comparison of the Flair model~\citep{stewart2022mwo2kg} and the \texttt{GPT-3.5} LLMs (non-fine-tuned and fine-tuned) on the test dataset. Support is the number of times the label appears in the test dataset. The results of the top-performing model (when there are no ties) are in \textbf{bold}.  \label{tab:table-fm-results}}
\end{table*}

\label{subsec:results2}
We now aim to determine whether fine-tuning the LLM on a purpose-built dataset is necessary, or whether similar performance can be achieved without fine-tuning. We focus our attention on a comparison between the \texttt{GPT-3.5} model, and \texttt{GPT-3.5 (Fine-tuned)}. The former model has been fed with the prompt discussed at the end of \ref{subsec:results1}, i.e. it constrains the model to predict only the failure mode and nothing else, and also provides it with a list of the valid failure modes from the dataset. The latter model has been fine-tuned on the 500 (observation, label) pairs in the training dataset, and the prompt does not contain the aforementioned constraints (as they are not necessary due to the fine-tuning).

Table~\ref{tab:table-fm-results} shows the results of each model  on the test dataset. It is clear that fine-tuning has a significant impact on performance, as the Micro-F1 score jumps from 0.46 to 0.81 between the non fine-tuned and fine-tuned models respectively. The results of the non fine-tuned model indicate that it does possess knowledge of maintenance, though, as it was capable of getting nearly half of all predictions correct without any form of fine-tuning. 

We also tested the effectiveness of ``few-shot learning'', i.e. providing a list of example (observation: failure mode) pairs to the model as part of the \texttt{system}-level prompt as opposed to a list of only the valid failure modes. We found that the results were near identical to the non fine-tuned model, and thus did not include these results in the table for brevity. Overall, the results show that fine-tuning is necessary to achieve strong performance. This demonstrates the importance of high quality annotated data when applying LLMs to maintenance work orders.


\subsection{Are LLMs more effective at failure mode classification than text classification models?}

To answer this final research question we focus our attention to a comparison between the \texttt{Flair} text classification model from~\citep{stewart2022mwo2kg} and the \texttt{GPT-3.5} models. As shown in Table~\ref{tab:table-fm-results}, the LLM significantly outperforms Flair, but only after fine-tuning. Without fine-tuning, Flair exhibits much stronger performance, indicating the necessity of annotated training data to be able to perform this particular task.

After fine-tuning on the annotated data, the LLM performs significantly better than Flair. It also tends to fair better on the minority classes, such as ``Failure to start on demand'', ``Failure to stop on demand'', etc, which we argue can be attributed to the underlying knowledge made available as part of the LLM's lengthy training process on a large corpora.

In summary, our results show this LLM is more effective at FMC than the text classification model, but only when the LLM is fine-tuned to perform this task.

\subsection{What are some barriers one may face when using LLMs for FMC?}

Overall we found the process of using and fine-tuning \texttt{GPT-3.5} fairly straightforward, though we experienced a couple of issues that are worth noting. Firstly, the non-deterministic nature of LLMs mean that they can produce different output given the same input. There is a built-in \texttt{temperature} parameter which can be set to \texttt{0} to reduce the likelihood of this occurring, but in our experience we were still receiving slightly different results each time we ran our experiments. This effect is most noticeable in the non fine-tuned model with no prompt engineering (i.e. from Section~\ref{subsec:results1}, and has less of an effect when the model is informed of the list of valid labels.

We also noticed that during inference, the OpenAI API would occasionally refuse our requests due to being overloaded, causing us to have to start the inference process again. This was not a significant problem for our small 62-record test set, but it would be more problematic when running inference over a large dataset.

Finally, we note that the overall fine-tuning and inference process was fairly inexpensive, costing approximately \$1 USD for each of our experiments. This shows that cost is not a barrier for achieving an acceptable level of performance on failure mode classification using LLMs.


%

\section{Conclusion}

In this paper we have demonstrated the use of Large Language Models (LLMs) to perform Failure Mode Classification (FMC). We have investigated the use of prompt engineering to determine the best prompt to feed in to an LLM, such as \texttt{GPT-3.5}, in order to perform FMC without any fine-tuning. However, we have also found that fine-tuning an LLM is necessary to obtain significantly better performance on FMC when compared to text classification models such as Flair. The fine tuning is performed using a relatively small, high quality, annotated data set.

The annotated data set we used for fine-tuning is publicly available. It maps observations to failure modes based on ISO 14224 classes. For the benefit of industry users wishing to use this fine-tuned data set on their own data, we note they will need to preprocess their maintenance work orders to extract observations. An example of a code pipeline to do this is in \citep{stewart2022mwo2kg}. 

One of the key drawbacks of OpenAI's LLMs is that to be able to fine-tune the models, one must upload potentially sensitive data to OpenAI's servers. This is a non-issue for companies with the capability to run and fine tune LLMs in their own secure environments, but presents complications for others. In light of this, in the future we aim to investigate the performance of offline large language models, such as LLaMA~\citep{touvron2023llama}, on  failure mode classification. We also plan to explore how well the Flair-based model performs on this task when it is fed with GPT-based embeddings. Finally, we also plan to release a larger annotated dataset than the one proposed by~\citep{stewart2022mwo2kg}, which will enable further fine-tuning and improved evaluation quality.


\subsubsection*{Acknowledgments} This research is supported by the Australian Research Council through the Centre for Transforming Maintenance through Data Science (grant number IC180100030), funded by the Australian Government.

\bibliographystyle{unsrtnat}





\end{document}